\documentclass[]{config/antgroup}
\usepackage{config/antgroup}

\newcommand*\justify{%
  \fontdimen2\font=0.4em
  \fontdimen3\font=0.2em
  \fontdimen4\font=0.1em
  \fontdimen7\font=0.1em
  \hyphenchar\font=`\-
}

\renewcommand{\texttt}[1]{%
\begingroup
\ttfamily
\begingroup\lccode`~=`/\lowercase{\endgroup\def~}{/\discretionary{}{}{}}%
\begingroup\lccode`~=`[\lowercase{\endgroup\def~}{[\discretionary{}{}{}}%
\begingroup\lccode`~=`.\lowercase{\endgroup\def~}{.\discretionary{}{}{}}%
\catcode`/=\active\catcode`[=\active\catcode`.=\active
\justify\scantokens{#1\noexpand}%
\endgroup
}

\PassOptionsToPackage{numbers, compress}{natbib}

\usepackage[utf8]{inputenc}         %
\usepackage[T1]{fontenc}            %
\usepackage{graphicx}
\usepackage{todonotes}
\usepackage{multirow}
\usepackage{hyperref}
\usepackage[capitalize,noabbrev,nameinlink,sort]{cleveref}
\usepackage{subcaption}
\usepackage{multicol}
\usepackage{array}
\usepackage{enumitem}
\usepackage{algorithm}
\usepackage{algpseudocode}
\usepackage{tabularx}
\usepackage{listings}
\usepackage{makecell}
\usepackage{xspace}
\usepackage{colortbl}
\usepackage{fontawesome5}
\usepackage{pifont}
\usepackage[normalem]{ulem}
\useunder{\uline}{\ul}{}
\usepackage{tablefootnote}
\usepackage{tcolorbox}
\usepackage{arydshln}
\usepackage[toc, page, header]{appendix}
\usepackage{tabularx}
\usepackage{listings}

\usepackage{amsmath}
\usepackage{amssymb}
\usepackage{amsfonts}                               
\usepackage{amsthm}
\usepackage[mathcal]{eucal}
\usepackage{mathrsfs}
\usepackage{bm}                                     
\usepackage{blkarray}                               
\usepackage{nicefrac}                               
\usepackage{bbm}

\usepackage{wrapfig}
\usepackage{graphicx}                               
\usepackage{caption}
\usepackage{tikz}
\usepackage{circuitikz}
\usetikzlibrary{patterns,snakes}
\usetikzlibrary{positioning,calc,fit,decorations.pathmorphing,shapes.geometric, shapes.gates.logic.US, calc}
\usetikzlibrary{arrows,arrows.meta,decorations.markings,shapes,shapes.arrows}
\usetikzlibrary{decorations,decorations.pathreplacing}
\usetikzlibrary{backgrounds}
\usepackage{filecontents}                           
\usepackage{pgfplots}
\usepackage{pgfplotstable}
\usepgfplotslibrary{groupplots}
\usepackage{scalefnt}
\pgfplotsset{compat=newest}
\usepackage{xcolor}

\definecolor{firstcolor}{HTML}{C3423F}
\definecolor{secondcolor}{HTML}{2A4B8C}
\definecolor{aworld_blue}{HTML}{4e81ff}
\definecolor{aworld_cyan}{HTML}{41d7fa}
\definecolor{aworld_teal}{HTML}{5fede4}
\definecolor{coral}{RGB}{255,127,80}
\definecolor{darkgreen}{RGB}{0,100,0}
\definecolor{darkyellow}{RGB}{204,153,0}
\definecolor{salmon}{RGB}{250,128,114}
\definecolor{darkred}{RGB}{150,0,0}

\hypersetup{
  colorlinks=true,
  linkbordercolor=aworld_blue,
  citebordercolor=aworld_cyan,
  urlbordercolor=aworld_teal
}

\definecolor{improvementblue}{RGB}{55,126,184}    
\definecolor{degradationorange}{RGB}{230,85,13}   

\def\eqref#1{equation~\ref{#1}}

\def\1{\bm{1}}

\DeclareMathAlphabet{\mathsfit}{\encodingdefault}{\sfdefault}{m}{sl}
\SetMathAlphabet{\mathsfit}{bold}{\encodingdefault}{\sfdefault}{bx}{n}

\begin{document}


\title{FunReason-MT Technical Report: Advanced Data Synthesis Solution for Real-world Multi-Turn Tool-use}

\author{
   Zengzhuang Xu$^{1\dagger\ast}$, 
   Bingguang Hao$^{1\dagger\ast}$, 
   Zechuan Wang$^{2\ast}$,
   Yuntao Wen$^{1\ast}$,
   Xinyi Xu$^{1\ast}$,
   Yang Liu$^{2\ast}$,
    \\
   \vspace{0.5em}
   Long Chen$^{1\ast}$,
   Dong Wang$^{1\ast}$,
   Maolin Wang$^{3\ast}$,
   Tong Zhao$^{2\ast}$,
  Yicheng Chen$^{1\ast}$,
   Cunyin Peng$^{1\ast}$,
      \\
   \vspace{0.5em}
   Jinjie Gu$^1$,
   Leilei Gan$^2$,
   Xiangyu Zhao$^3$,
  Chenyi Zhuang$^{1\ddagger}$,
   Shi Gu$^{2\ddagger}$
 }

{\renewcommand\thefootnote{}
\footnotetext{$^\dagger$Equal Contribution. The two authors designed the system.}
\footnotetext{$^\ast$Core Contributor. These authors participated in algorithm development, data engineering, and infrastructure.}
\footnotetext{$^\ddagger$Corresponding Author. chenyi.zcy@antgroup.com, gus@zju.edu.cn}
\footnotetext{Work was done during Bingguang’s internship at Ant Group. bingguanghao7@gmail.com}
}

\affiliation{\raisebox{-0.2em}{\includegraphics[height=1.1em]{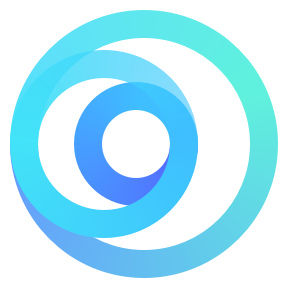}}$^1$AWorld Team, Inclusion AI\quad}
\affiliation{$^2$Zhejiang University\quad}
\affiliation{$^3$City University of Hong Kong\quad}

\maketitle


\begin{center}
  \vspace{-2em}
\href{https://huggingface.co/datasets/Bingguang/FunReason-MT}{\raisebox{-0.2em}{\includegraphics[height=1.1em]{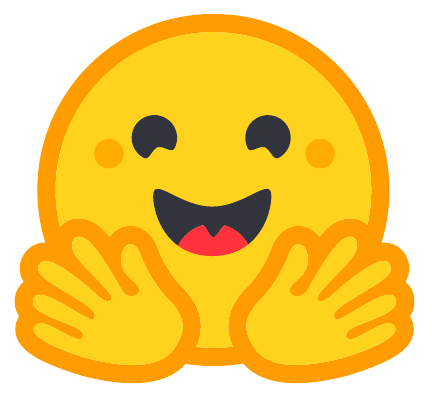}} \texttt{FunReason-MT Dataset}}\quad\quad
\href{https://huggingface.co/Bingguang/FunReason-MT}{\raisebox{-0.2em}{\includegraphics[height=1.1em]{logo/hf-logo.pdf}} \texttt{FunReason-MT Model}}\quad\quad\href{https://github.com/inclusionAI/AWorld-RL}{\faGithub\ \texttt{Project FunReason-MT}}
\end{center}


\begin{abstract}
Function calling (FC) empowers large language models (LLMs) and autonomous agents to interface with external tools, a critical capability for solving complex, real-world problems. As this ability becomes increasingly central to advanced AI systems, the need for high-quality, multi-turn training data to develop and refine it cannot be overstated. Existing data synthesis methods, such as random environment sampling or multi-agent role-playing, are not powerful enough to generate high-quality data in real-world environments. Practical challenges come in three folds: targeted data synthesis, hard query construction, and multi-turn logical dependency. To address these structural deficiencies, we present FunReason-MT, a novel data synthesis framework for real-world multi-turn tool use. FunReason-MT resolves the complexity barrier in multi-turn FC data by employing 1) \emph{Environment-API Graph Interactions} to gather varied high-quality trajectories with targeted tool, 2) \emph{Advanced Tool-Query Synthesis} to simplify hard query construction, and 3) \emph{Guided Iterative Chain} for sophisticated CoT generation. Evaluations on Berkeley Function-Calling Leaderboard (BFCLv3) demonstrate the power of our framework: a 4B model built upon FunReason-MT generated data achieves state-of-the-art performance among comparable-sized models. Further performance improvements on BFCLv4 confirm that FunReason-MT provides a reliable and robust source for agentic learning. 
\\
\\
\textbf{This is part of the open-source Project \raisebox{-0.2em}{\includegraphics[height=1.1em]{logo/aworld_logo.png}}\href{https://github.com/inclusionAI/AWorld}{\texttt{AWorld}}, Inclusion AI. The training data and model weights are released.}
\end{abstract}

\begin{figure*}[!h]
    \centering
    \begin{subfigure}{0.3\textwidth}
        \centering
        \includegraphics[width=\textwidth]{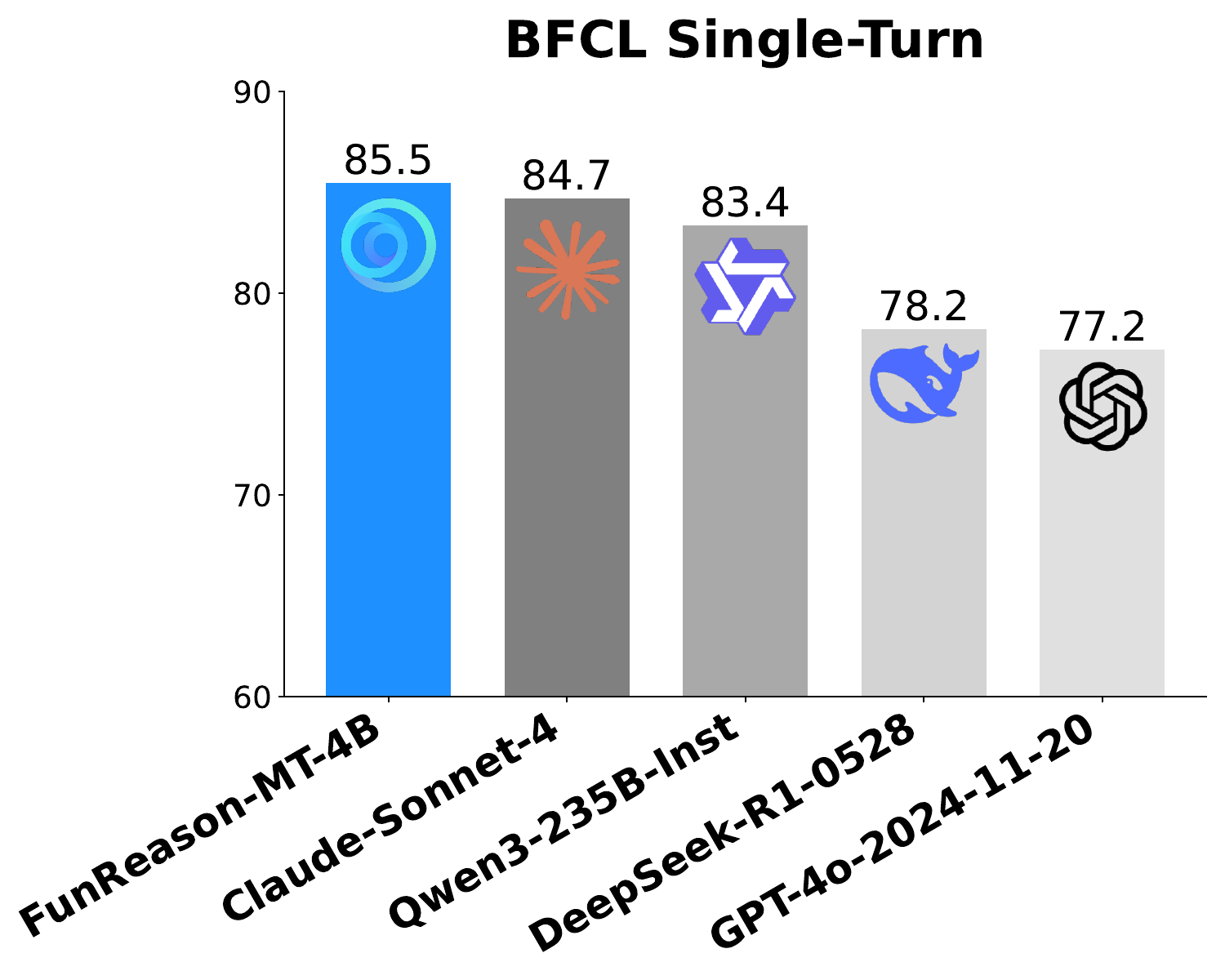}
    \end{subfigure}
    \begin{subfigure}{0.3\textwidth}
        \centering
        \includegraphics[width=\textwidth]{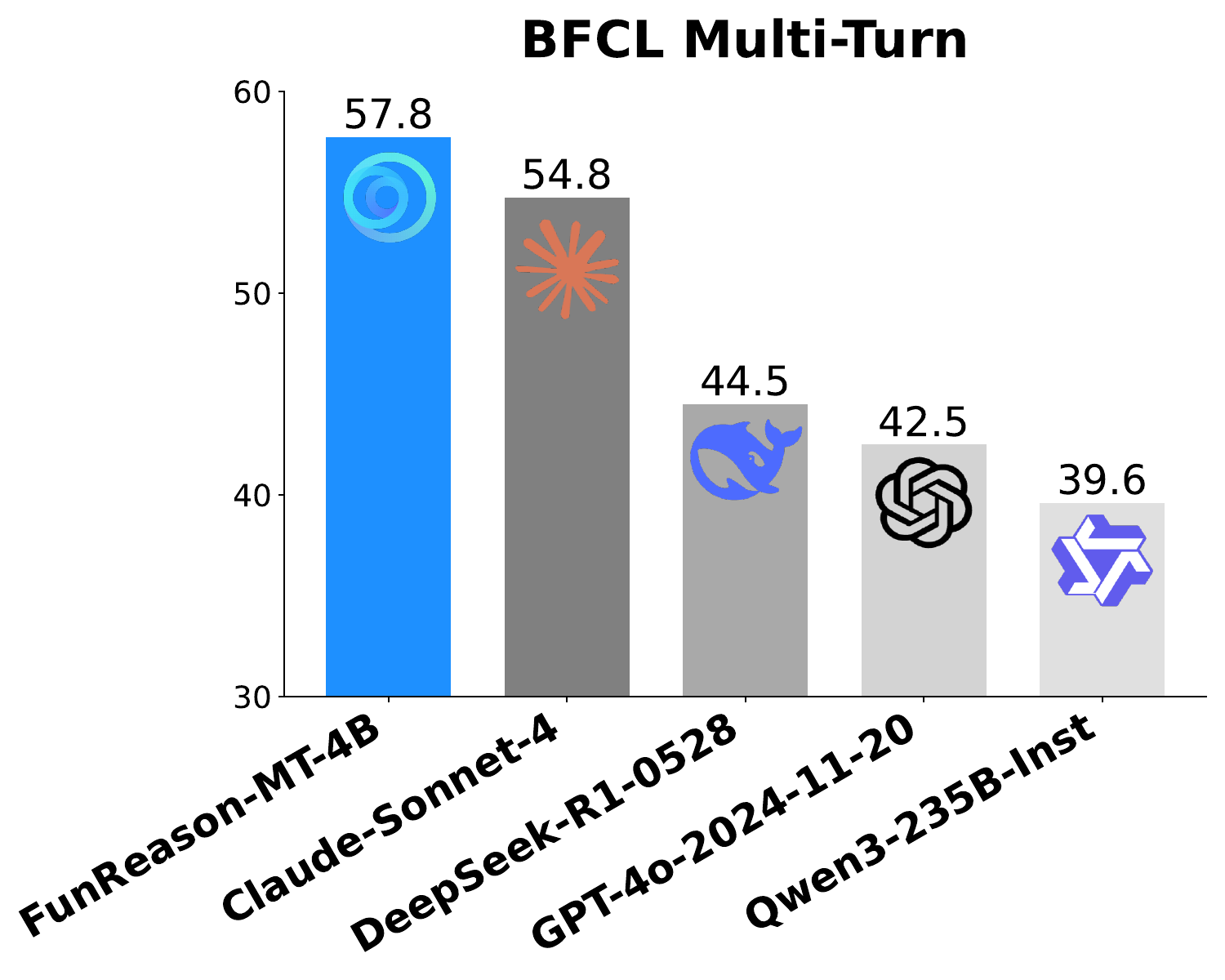}
    \end{subfigure}
    \begin{subfigure}{0.3\textwidth}
        \centering
        \includegraphics[width=\textwidth]{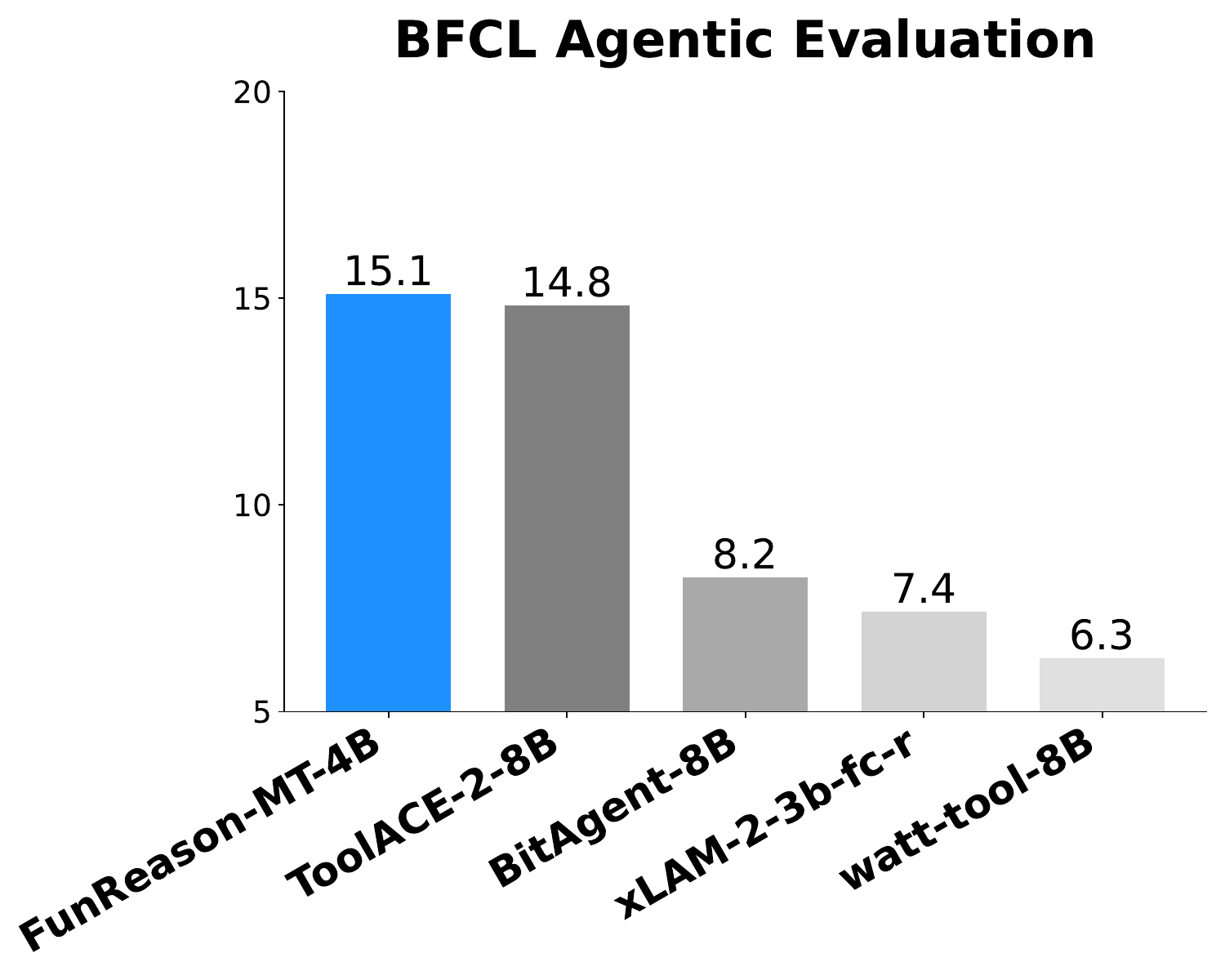}
    \end{subfigure}
    \caption{Performance on BFCL Single-Turn, Multi-Turn and Agentic Evaluation.}
    \label{tab:model_performance}
\end{figure*}

\newpage

\section{Introduction}
\label{sec:intro}

\begin{figure}
    \centering
    \includegraphics[width=1.0\textwidth]{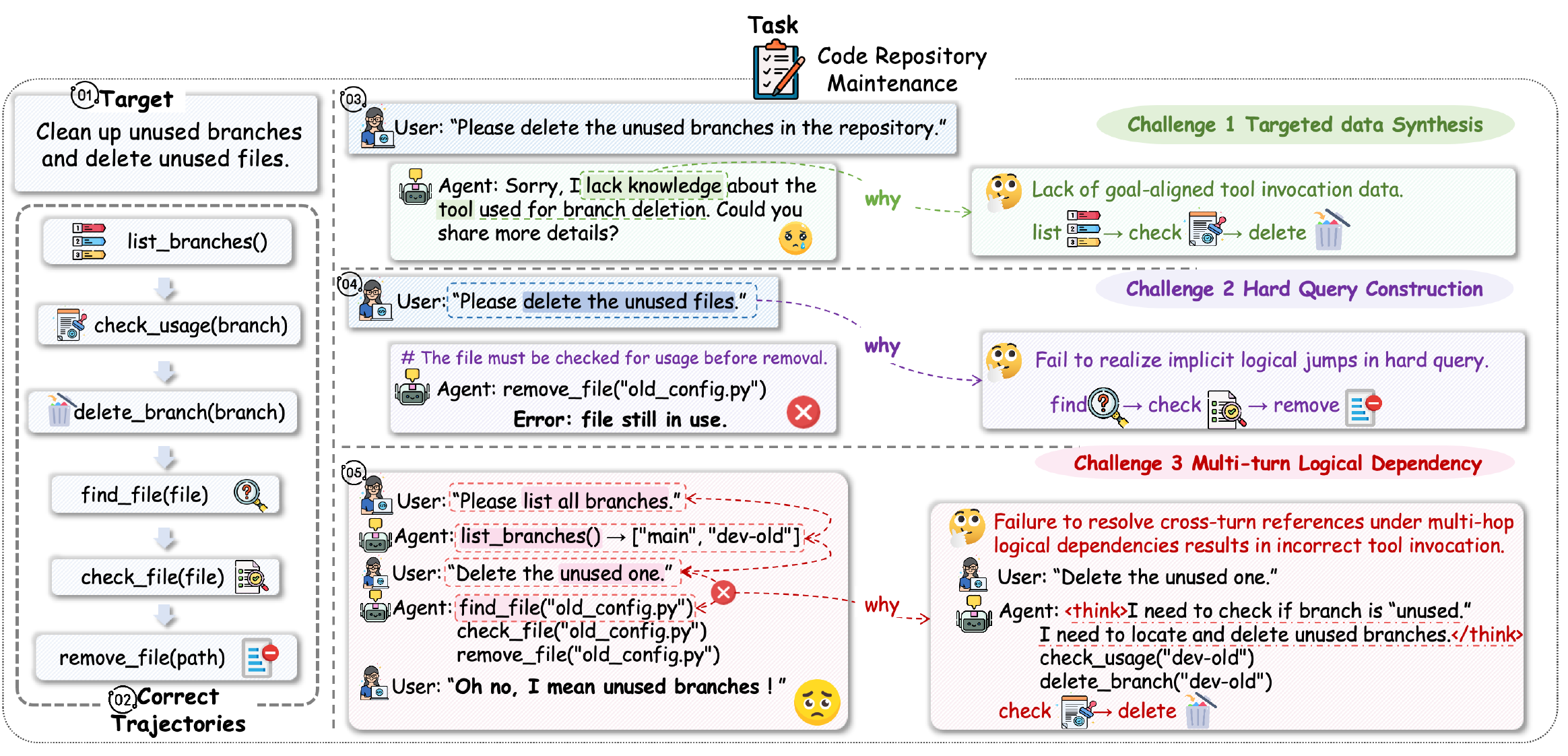}
    \caption{We summarize the core challenges in multi-turn function-calling data generation into three aspects.}
    \label{fig:challenge}
\end{figure}

Large Language Models (LLMs) are profoundly transforming human-computer interaction, largely through Function Calling (FC), the pivotal technology enabling LLMs to interface with external tools~\citep{wang2025function}. Although contemporary foundation models possess inherent tool-use capability, unlocking more advanced applications in this domain is hindered by a critical bottleneck: The construction of effective, high-quality data, particularly for multi-turn trajectories. Existing data generation research predominantly relies on constrained methodologies, such as random sampling within environments or Multi-Agent System (MAS) role-playing simulations~\citep{huang2025ttpa,prabhakar2025apigen,liu2024apigen,zeng2025toolace}.
Crucially, random sampling fails to capture rare, complex events, while MAS role-playing often defaults to simplistic ``happy-path'' scenarios or suffers from a lack of true diversity~\citep{han2024llm}. Thus these approaches inherently produce data that is limited in both effectiveness and diversity, resulting in a substantial training performance ceiling.

As illustrated in Figure~\ref{fig:challenge}, we categorize the challenge in multi-turn FC data generation into three critical, interconnected issues that diminish data quality and model stability. \textbf{1) Targeted Data Synthesis.} Existing random sampling techniques fail to controllably and targetedly construct trajectories requiring the collaborative use of a target complex tool with others. \textbf{2) Hard Query Construction.} Due to the modular inputs of individual tools rather than holistic integration, direct environmental sampling and MAS role-playing struggle to synthesize hard, logical-jump queries. \textbf{3) Multi-Turn Logical Dependency.} In complex multi-turn dialogues, each step of the Chain-of-Thought (CoT) is a validation and reconstruction of the reliance on prior logic. Current Reasoning Large Language Models (RLLMs) often falter when generating CoT in unexplored environments, preventing the acquisition of complete and accurate multi-turn trajectories.

To fundamentally resolve these structural limitations, we introduce the \textbf{Fun}ction Call \textbf{Reason}ing \textbf{M}ulti-\textbf{T}urn (\textbf{FunReason-MT}) data synthesis framework. Our framework prioritizes robustness and efficiency, and incorporates three specific core components to tackle the outlined challenges. 

\begin{itemize}
\item \textbf{Environment-API Graph Interactions.} We establish an API graph based on interdependencies between tools and environments, enabling the collection of multi-turn trajectories centered around a targeted complex API within varied scenarios, ultimately training models for targeted tool mastery and collaborative use.

\item \textbf{Advanced Tool-Query Synthesis.} We use the tool-state pairs collected during Environment-API Graph Interactions to construct an advanced tool that eliminates the need to input the modularized sub-tools when queries are generated in reverse. This inherently makes the execution of prerequisite tools implicit.

\item \textbf{Guided Iterative Chain.}  We input the hard query, the set of sub-tools, and the advanced tool description to obtain CoT. We then implement an iterative feedback loop: The analysis identifying the specific reasons for any failure to match the ground truth is continuously integrated as a corrective hint until the CoT successfully leads to the right FC or meets the maximum attempts.

\end{itemize}

We validate the FunReason-MT framework on the challenging Berkeley Function-Calling Leaderboard (BFCLv3), encompassing both single-turn and multi-turn tests~\citep{patilberkeley}. Experiments demonstrate that our data construction method consistently outperforms approaches based on MAS role-playing or random environmental sampling, and achieves state-of-the-art performance among comparable-sized models. This success affirms that FunReason-MT effectively overcomes the complexity barrier in multi-turn FC data generation, establishing an essential knowledge supplement for future LLM environment exploration. We also evaluate our models on BFCLv4, an out-of-distribution (OOD) benchmark, to further assess their agentic ability. Although our data is unrelated to the tasks evaluated by BFCLv4 (Web Search and Memory), our models still show improvement in these specific areas. We believe our study is well-positioned to lay a robust and enduring foundation that facilitates the future progression of agentic learning~\citep{zhang2025landscape,ke2025survey}.

\section{Motivation}
\label{sec:motivation}

Our motivation stems from the insight that the identified ``performance ceiling'' is not merely a data quantity issue, but a fundamental failure of the existing data generation paradigm. We recognized that ``bottom-up'' approaches like random sampling and MAS role-playing are structurally incapable of producing the high-complexity, long-tail trajectories needed for true mastery. These methods merely hope for complexity to emerge from simple interactions, suffer from low controllability and poor reliability. 

This drove us to devise a new, ``top-down'' construction methodology that can explicitly direct the model to master a specific complex tool within a targeted complex scenario. Our approach is motivated by the ``why'' and ``how'' of this shift: instead of hoping for targeted complex tool involved trajectories, we engineer them by modeling the environment's structural dependencies (the API graph). Instead of sampling simple tool combinations, we guarantee complex, logical-jump queries by generating them in reverse. And finally, instead of accepting flawed reasoning, we enforce CoT accuracy through a guided feedback loop, thus directly targeting the three core challenges that plague existing methods and corresponding to solving the issues of \textbf{reliability} and \textbf{complexity} respectively.


\section{Notation}

In order to describe the generation process of the FunReason-MT framework, we introduce the key symbols and concepts used to model the system's components and operations. The core concepts and models that manage the system are defined in Table~\ref{tab:coresets}.

\begin{table}[h]
\centering
\begin{tabular}{ll}
\toprule
\textbf{Symbol} & \textbf{Concept} \\
\midrule
$\mathcal{S}$ & Multi-Environment Simulation Space. \\
$\mathcal{T}$ & Tool Set. \\
$\mathcal{G} = (\mathcal{T}, \mathcal{R}, \mathcal{P})$ & API Relation Graph (Tools, Dependencies $\mathcal{R}$, Parameters $\mathcal{P}$). \\
$A_T, A_Q, A_R, A_C$ & LLM Agents (Tooling, Querying, Reasoning, Critiquing). \\
$T_a \in \mathcal{T}$ & Target Tool. \\
$M$ & Number of tool calls per round. \\
$N$ & Number of trajectory rounds. \\
$K_{\max}$ & Maximum self-correction attempts. \\
$\text{Traj}$ & Multi-turn trajectory. \\
$\text{Turn}_i$ & i-th single-turn content. \\
\bottomrule
\end{tabular}
\caption{Core Concepts and Models of FunReason-MT.}
\label{tab:coresets}
\end{table}

\section{Methodology}
\label{sec:methodology}

\begin{figure}[t]
    \centering
    \includegraphics[width=1.0\linewidth]{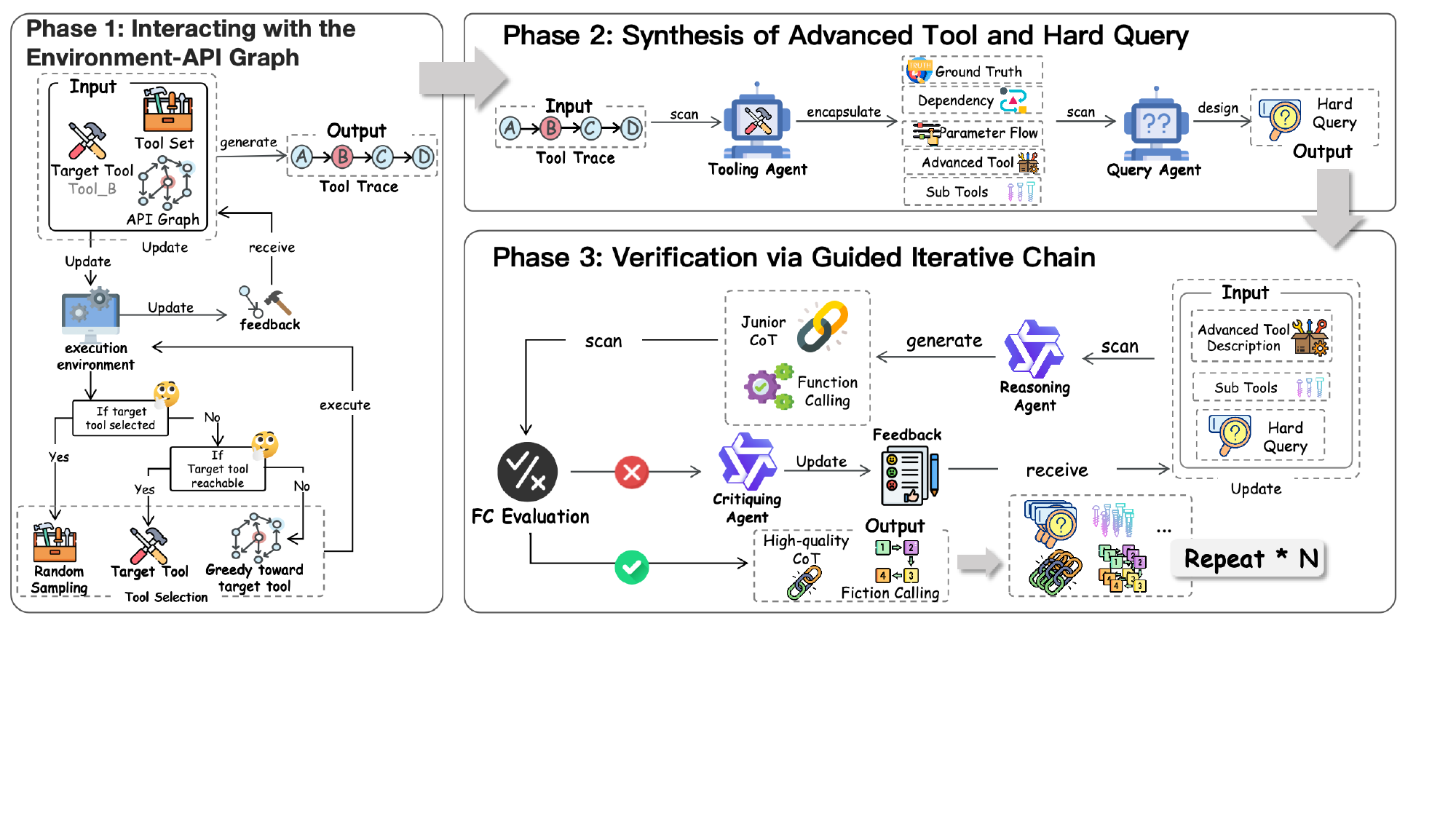}
    \caption{Three Phases of FunReason-MT.}
    \label{fig:pipeline}
\end{figure}

As shown in Figure~\ref{fig:pipeline}, our proposed FunReason-MT framework comprises three core phases:
(1) \textbf{Phase I: Environment-API Graph Interactions} constructs a valid multi-step execution trace by sampling tool calls from the API Relation Graph (\S\ref{sec:phase1});
(2) \textbf{Phase II: Advanced Tool-Query Synthesis} reverse-engineers a hard-to-solve data sample and synthesizing a corresponding challenging Hard Query ($Q_{\text{hard}}$) (\S\ref{sec:phase2});
(3) \textbf{Phase III: Guided Iterative Chain} utilizes an iterative, self-correction loop to refine the CoT and ensure its logical consistency with the ground truth answer (\S\ref{sec:phase3}).

\subsection{Phase I: Environment-API Graph Interactions}
\label{sec:phase1}
A central challenge in generating multi-turn tool-use trajectories is to devise a sampling strategy that is both \textbf{executable} and \textbf{purposefully directed}. To achieve this, FunReason-MT framework is designed to satisfy two critical constraints simultaneously: \textit{execution correctness} (\emph{i.e.}, respecting tool dependencies) and \textit{goal-directed efficiency} (\emph{i.e.}, prioritizing $T_a$). We detail these two components below.

\subsubsection*{Formalizing Tool Call Legality}

To ensure a valid execution sequence, we introduce a strict dependency constraint: Tool $T_i$ is only callable if all its necessary preconditions, $\text{Prerequisite}(T_i)$, are satisfied by the set of already executed tools, $\mathcal{T}_{\text{called}}$. We formalize this essential constraint using the tool legality check $I$:
\begin{equation}
I(T_i, \mathcal{T}_{\text{called}}) = \mathbf{1}_{\{\text{Prerequisite}(T_i) \subseteq \mathcal{T}_{\text{called}}\}},
\end{equation}
where $\mathbf{1}_{\{\cdot\}}$ is the indicator function, returning 1 if the condition is true and 0 otherwise.

\subsubsection*{Tool Sampling with Priority}

While random sampling over the tool graph can satisfy execution correctness, it often fails to reach the target tool $T_a$ efficiently.
To address this, we introduce a Directed Sampler, denoted as $\text{Sampler}$, which explicitly biases the sampling process toward $T_a$.
Specifically, $\text{Sampler}$ adopts a greedy heuristic strategy that operates over the set of legal tools $\mathcal{T}_{\text{legal}}$ and prioritizes those minimizing the graph distance $\text{dist}(\cdot,\cdot)$ to $T_a$. The sampling policy of $\text{Sampler}$ is formally defined as follows:

\begin{equation}
\label{eq:tool_sampling}
T_s = \text{Sampler}(T_a, \mathcal{T}_{\text{called}}) = \begin{cases}
\text{rand}(\mathcal{T}_{\text{legal}}), & \text{if } T_a\in \mathcal{T}_{\text{called}},\\
T_a, & \text{if } I(T_a, \mathcal{T}_{\text{called}}) = 1\land T_a \notin \mathcal{T}_{\text{called}},\\
\underset{T_k}{\operatorname{argmin}} \left\{ \text{dist}(T_k, T_a) \right\}, & \text{otherwise}.
\end{cases}
\end{equation}
The chosen tool $T_s$ is executed using sampled parameters $P_s$ as the call $C_s = (T_s, P_s)$. This action yields environmental feedback $E_i$, triggers an update to the system state $\mathcal{S}_j$, and consequently updates the API Graph $\mathcal{G}$, legal tools $\mathcal{T}_{\text{legal}}$ and executed tools $\mathcal{T}_{\text{called}}$. The resultant single-turn execution sequence is recorded as the trace: $\text{Turn}_i = (C_1,E_1,C_2, E_2, \dots,C_M, E_M)$.

\subsection{Phase II: Advanced Tool-Query Synthesis}
\label{sec:phase2}

The execution trace $\text{Turn}_i$ ensures correct tool execution, yet it fails to capture the abstract intent behind those operations. So we need a mechanism that encourages the model to reinterpret the entire trace as a single high-level abstraction. To realize this objective, Phase~II first abstracts the trace into an advanced tool and then constructs a hard query that depends on this abstraction.

\begin{enumerate}[leftmargin=*]

\item \textbf{Advanced Tool Generation.}
Given the execution trace $\text{Turn}_i$, the \emph{Tooling Agent} ($A_T$) synthesizes a single composite abstraction, denoted as the advanced tool $T_{\text{adv}}$:
\begin{equation}
T_{\text{adv}} = A_T\left(\text{Turn}_i\right),
\end{equation}
where $A_T$ abstracts the multi-step execution trace into a unified high-level operation that encapsulates the functionality and interdependencies of all sub-tools within $\text{Turn}_i$.

\item \textbf{Hard Query Construction.}
Conditioned on the synthesized advanced tool $T_{\text{adv}}$, the \emph{Querying Agent} ($A_Q$) generates a challenging hard query $Q_{\text{hard}}$ that explicitly requires the use of $T_{\text{adv}}$ for resolution.  
Formally, the construction process is expressed as:
\begin{equation}
Q_{\text{hard}} = A_Q\left(T_{\text{adv}}, \epsilon\right),
\end{equation}
where $\epsilon$ denotes a synthesis noise that increases task difficulty and encourages generalization.

\end{enumerate}

\subsection{Phase III: Guided Iterative Chain}
\label{sec:phase3}

The construction of $Q_{\text{hard}}$ promotes higher-level reasoning, but it also increases the risk of logical inconsistency or incomplete Chain-of-Thought (CoT) generation.
To mitigate these issues, Phase~III introduces the Guided Iterative Chain, a feedback-driven refinement process that iteratively improves CoT quality and coherence until the reasoning converges to a reliable solution.

\subsubsection*{Initial Reasoning Attempt and Correction Loop}
The \emph{Reasoning Agent} ($A_R$) begins by attempting to resolve the hard query, $Q_{\text{hard}}$, utilizing both the advanced tool description $\text{Des}(T_{\text{adv}})$ (denoted as $\text{Prompt}^{(1)}$ in the first attempt) and its primitive components $\mathcal{T}_{\text{sub}}$. 

The $k^{th}$  attempt yields the output $\{\text{FC}^{(k)}, \text{CoT}^{(k)}\} = A_R(Q_{\text{hard}}, \mathcal{T}_{\text{sub}},  \text{Prompt}^{(k)})$, comprising a CoT and a Function Call. The generated function call ($\text{FC}^{(k)}$) is subsequently validated against the known ground truth $G$ ( $G=(C_1, C_2, \dots, C_M)$ ) using a verification function $\text{Validate}: \text{FC}^{(k)} \times G \to \{\text{Pass}, \text{Fail}\}$.

\begin{enumerate}[leftmargin=*]
\item \textbf{Error Analysis (Upon Failure):}  If the current attempt fails, the \emph{Critiquing Agent} ($A_C$) analyzes the incorrect function call ($\text{FC}^{(k)}$) with respect to the ground truth $G$, identifies the specific error, and generates targeted corrective hint feedback: $\text{Error}^{(k)} = A_C(\text{FC}^{(k)}, G)$.

\item \textbf{Self-Correction:} In this step, the \emph{Reasoning Agent} ($A_R$) first updates its context using the corrective hint $\text{Error}^{(k)}$ from the previous turn, forming an augmented prompt that explicitly conditions the next attempt on the identified error: $\text{Prompt}^{(k+1)} = \text{Concat}(\text{Prompt}^{(k)}, \text{Error}^{(k)})$. Given this augmented prompt, $A_R$ is then re-invoked to generate a refined solution, $\{\text{FC}^{(k+1)}, \text{CoT}^{(k+1)}\} = A_R(Q_{\text{hard}}, \mathcal{T}_{\text{sub}}, \text{Prompt}^{(k+1)})$. This self-correction step, together with error analysis, induces an iterative feedback loop that progressively improves subsequent attempts.

\end{enumerate}

The maximum allowed limit for this iterative correction process is $K_{\max}$ attempts. A data sample is only retained if the validation function  $\text{Validate}(\text{FC}^{(k)}, G)$ returns $\text{Pass}$ before the iteration cap is reached.

\subsection{Scaling from Single-Turn Trace to Multi-Turn Trajectory}

The full pipeline is iterated $N$ times to scale the single-turn generation into a comprehensive, high-quality multi-turn trajectory, $\text{Traj}$.

The overall data generation process starts by initializing $\text{Traj}$ as an empty set. The framework then sequentially executes the three phases for each of the $N$ required data rounds:

\begin{enumerate}[label=\textbf{Step \arabic*:}, leftmargin=*, align=left, itemsep=5pt]
\item \textbf{Environment API-Graph Interactions:} A target tool $T_a$ is chosen from the primitive tool set $\mathcal{T}$. This choice dictates the generation of a complete execution trace $\text{Turn}_1$ within the simulation environment, following the procedure outlined in Phase 1.
\item \textbf{Advanced Tool-Query Synthesis:} The successful trace is then gave to the LLM agents $A_T$ and $A_Q$. These agents collaborate to synthesize the Advanced Tool ($T_{\text{adv}}$) and its corresponding Hard Query ($Q_{\text{hard}}$), thereby forming the core of the new data sample.
\item \textbf{Guided Iterative Chain:} Immediately following synthesis, the data sample undergoes the crucial Guided Iterative Chain verification process. This step employs the reasoning and critiquing agents ($A_R, A_C$) to attempt solving the hard query and validate results against the ground truth answer $G$.
\end{enumerate}

To ensure the resulting multi-turn trajectory is reliable and complex, the entire process detailed above will be repeated $N$ times.
\section{Experiments}
\label{sec:experiments}

\subsection{Experimental Setup}
\label{subsec:exp_setup}

\textbf{Training Details.} Considering the length of multi-turn conversation and CoT, we use Qwen3-4B-Instruct-2507 with 256K long-context~\citep{yang2025qwen3} as our base model. The collected trajectories are split at each assistant response and the model is only trained on the CoT and answer at current turn. We collect 17000 multi-turn samples, and to enhance the diversity of the dataset, we also jointly train model with function calling data from APIGen. We use Llama-Factory~\citep{zheng2024llamafactory} and Verl~\citep{sheng2024hybridflow} library to train models with implementations from~\cite{hao2025funreason} and~\cite{hao2025exploring}.

\textbf{Benchmarks.} Our primary evaluation is performed on the BFCLv3, which is designed to assess the model's performance on both Single-Turn and Multi-Turn tool-calling tasks. To further evaluate the data influence on the model's agentic capability, we extend our assessment to the \textbf{Held-out (OOD)} BFCLv4 benchmark. BFCLv4 is engineered for the evaluation of agentic performance with Web Search and Memory. We present all the results in terms of accuracy (\%).

\subsection{Main Results}
\label{subsec:main_results}
\begin{table}[!htp]
\small
\centering
\begin{tabular}{@{}lcccccccc@{}}
\toprule
& \multicolumn{5}{c}{\textbf{Multi-Turn}} & \multicolumn{3}{c}{\textbf{Single-Turn}}\\
\cmidrule(lr){2-6}\cmidrule(lr){7-9}
\textbf{Model}
& \textit{Overall} 
& \textit{Base} 
& \begin{tabular}[c]{@{}c@{}}\textit{Miss}\\ \textit{Func}\end{tabular}
& \begin{tabular}[c]{@{}c@{}}\textit{Miss}\\ \textit{Param}\end{tabular}
& \begin{tabular}[c]{@{}c@{}}\textit{Long}\\ \textit{Context}\end{tabular}
& \textit{Overall} 
& \textit{Non-Live} 
& \textit{Live} \\ 
\midrule
\rowcolor{gray!15}
\multicolumn{9}{l}{\emph{\textbf{Close Source Model}}} \\[1pt]
GPT-5-2025-08-07 & 28.50 & 33.50 & 29.50 & 23.00 & 28.00 & 65.59 & 72.92 & 58.25\\
GPT-4o-2024-11-20 & 42.50 & 55.50 & 34.50 & 29.00 & 51.00 & 77.21 & 83.88 & 70.54\\
Claude-Sonnet-4-20250514 & 54.75 & \underline{64.00} & 54.00 & 47.50 & \underline{53.50} & \underline{84.72} & 88.38 & \underline{81.05}\\
Gemini-2.5-Pro & 25.00 & 25.50 & 26.00 & 24.50 & 24.00 & 74.50 & 85.04 & 63.95\\
o3-2025-04-16 & 38.38 & 44.00 & 40.50 & 31.50 & 37.50 & 53.01 & 39.98 & 66.03\\
Grok-4-0709 & 36.12 & 44.00 & 31.00 & 26.00 & 43.50 & 79.80 & 85.21 & 74.39\\
\midrule
\rowcolor{gray!15}
\multicolumn{9}{l}{\emph{\textbf{Open Source Model}}} \\[1pt]
Moonshotai-Kimi-K2-Inst & 41.25 & 51.00 & 43.00 & 31.00 & 40.00 & 80.80 & 84.02 & 77.57\\
DeepSeek-R1-0528 & 44.50 & 54.50 & 41.00 & 36.50 & 46.00 & 78.22 & 75.73 & 80.90\\
Qwen3-235B-A22B-Inst-2507 & 39.62 & 53.50 & 34.50 & 27.50 & 43.00 & 83.37 & \textbf{90.12} & 76.61\\
ToolACE-2-8B & 37.00 & 47.00 & 31.00 & 28.00 & 42.00 & 82.54 & 87.87 & 77.20\\
BitAgent-8B & 37.75 & 46.50 & 37.50 & 24.00 & 43.00 & 81.71 & 87.33 & 76.09\\
watt-tool-8B & 37.88 & 45.50 & 39.00 & 24.00 & 43.00 & 81.71 & 87.54 & 75.87\\
xLAM-2-3b-fc-r & \underline{57.12} & \textbf{73.50} & \underline{55.00} & \textbf{54.50} & 45.50 & 73.82 & 83.31 & 64.32\\
ToolACE-MT & 40.25 & 57.50 & 31.50 & 34.00 & 38.00 & 78.23 & 84.94 & 71.52\\
\midrule
Qwen3-4B-Inst-2507 & 15.75 & 19.00 & 15.50 & 12.50 & 16.00 & 78.19 & 86.35 & 70.02\\
\rowcolor[HTML]{EAE8FD}+FunReason-MT (w/ SFT) & 46.90& 53.20 & 47.10 & 40.40 & 46.90 & 81.97& 83.36 & 80.57\\
{$\Delta$}& \textcolor[HTML]{006400}{+31.15}& \textcolor[HTML]{006400}{+34.20}& \textcolor[HTML]{006400}{+31.60}& \textcolor[HTML]{006400}{+27.90} & \textcolor[HTML]{006400}{+30.90}& \textcolor[HTML]{006400}{+3.78} & \textcolor[HTML]{8B0000}{-2.99}&\textcolor[HTML]{006400}{+10.55}\\
\midrule
\rowcolor[HTML]{EAE8FD}+FunReason-MT (w/ RL) & \textbf{57.75}&62.50 & \textbf{60.50} & \underline{48.50} & \textbf{59.50} & \textbf{85.47}& \underline{89.58} & \textbf{81.35}\\
{$\Delta$} & \textcolor[HTML]{006400}{+42.00}         & \textcolor[HTML]{006400}{+43.50}      & \textcolor[HTML]{006400}{+45.00}        & \textcolor[HTML]{006400}{+36.00}      & \textcolor[HTML]{006400}{+43.50}        & \textcolor[HTML]{006400}{+7.28}             & \textcolor[HTML]{006400}{+3.23}      & \textcolor[HTML]{006400}{+11.33}  \\
\bottomrule
\end{tabular}
\caption{\label{tab:bfcl-overall} Performance on BFCLv3 (last updated August 26, 2025). The best result within each category is highlighted in \textbf{bold}. The second best result is \underline{underlined}.}
\end{table}

As shown in Table~\ref{tab:bfcl-overall}, on the BFCLv3 benchmark, FunReason-MT yields notable improvements on Qwen3-4B-Instruct, raising the Multi-Turn score from 15.75 to 46.90 (+31.15) after SFT and to 57.75 (+42.00) after RL. Despite its 4B size, the FunReason-MT RL-trained model surpasses both strong open-source models (\emph{e.g.}, Kimi-K2-Inst, DeepSeek-R1) and leading close-source models (\emph{e.g.}, GPT-5, Gemini-2.5-Pro, Claude-Sonnet-4), achieving state-of-the-art results. In addition, FunReason-MT demonstrates balanced performance across all sub-metrics, indicating strong generalization and stability across reasoning dimensions.

Overall, these results demonstrate that FunReason-MT effectively enhances multi-turn reasoning and tool-use capabilities by utilizing high-quality synthesized data, resulting in consistent and reliable improvements for real-world function-calling tasks.

\subsection{Agentic Results}

To investigate the transference of agentic capabilities inherent in the FunReason-MT, we evaluate our models on the OOD BFCLv4 benchmark. Specifically, we assess models on Search and Memory subsets, showing the results in Table~\ref{tab:bfclv4}.

\begin{table*}[h]
\small

\centering
\begin{tabular}{@{}l c @{\hspace{2em}} c @{\hspace{2em}} c cc @{\hspace{2em}} ccc@{}}
\toprule
& &\multicolumn{3}{c}{\textbf{Web Search}}& \multicolumn{4}{c}{\textbf{Memory}}\\
\cmidrule(lr){3-5}\cmidrule(lr){6-9}
\textbf{Models}
&\textbf{Overall}& \textit{Overall} &
 \textit{Base} &
 \textit{No Snippet} &
 \textit{Overall} &
 \textit{KV} &
 \textit{Vector}&
 \textit{Recursive Sum} \\ 
 \midrule
ToolACE-2-8B&14.83 &9.00	&13.00	&5.00	&20.65	&7.74	&18.06	&36.13\\
BitAgent-8B	&8.24 &4.00	&4.00	&4.00	&12.47	&4.52	&20.65	&12.26\\
watt-tool-8B&6.30 &4.00	&7.00	&1.00	&8.60	&3.87	&13.55	&8.39\\
xLAM-2-3b-fc-r&7.42 &1.50	&1.00	&2.00	&13.33	&4.52	&10.97	&24.52\\
\midrule
Qwen3-4B-Instruct-2507 &8.85 &5.00	&4.00	&6.00	&12.69	&11.61	&11.61	&14.84\\

+FunReason-MT (SFT) & 9.99\rlap{\textcolor{green!60!black}{\scriptsize$_{+1.14}$}} & 9.00\rlap{\textcolor{green!60!black}{\scriptsize$_{+4.00}$}}  & 12.00 & 6.00 & 10.97\rlap{\textcolor{red!60!black}{\scriptsize$_{-1.72}$}} & 1.29 & 0.00 & 31.61\\
+FunReason-MT (RL) & 15.10\rlap{\textcolor{green!60!black}{\scriptsize$_{+6.25}$}} & 16.00\rlap{\textcolor{green!60!black}{\scriptsize$_{+11.00}$}} & 20.00 & 12.00 & 14.19\rlap{\textcolor{green!60!black}{\scriptsize$_{+1.50}$}} & 5.81 & 0.00 & 36.77\\
\bottomrule
\end{tabular}
\caption{\label{tab:bfclv4} Performance of on BFCLv4 (last updated August 26, 2025).}
\end{table*}

On the BFCLv4 benchmark, the FunReason-MT model demonstrates significant performance gains over its base model, Qwen3-4B-Instruct-2507. SFT increases the Overall performance score from 8.85 to 9.99 (+1.14), while RL further boosts it to 15.10 (+6.25). The RL model's strength is particularly evident in the Web Search Overall subset, where its score raises from 5.00 to 16.00 (+11.00), alongside an increase in the Memory Overall score from 12.69 to 14.19. Notably, the RL-trained FunReason-MT outperforms all other competing specialized models in the main Overall metric.

The FunReason-MT model, particularly after being trained with reinforcement learning, demonstrates significant performance improvements on the BFCLv4 benchmark. This proves that data generated by this paradigm consistently enhances the model's agentic capabilities. We believe this paradigm of procedural data generation process can lay a strong foundation and provide a robust initial knowledge base for future large-scale, environment-feedback based, agentic reinforcement learning.

\section{Conclusion}
\label{sec:conclusion}

In this technical report, we introduce FunReason-MT, a novel data synthesis framework designed to address critical bottlenecks in multi-turn function calling data generation. We identified that existing methods, such as random sampling and MAS role-playing, are insufficient, producing data that lacks diversity and fails to capture complex, logically-dependent trajectories. Our framework overcomes these limitations through three core components: Environment-API Graph Interactions for targeted data generation, Advanced Tool-Query Synthesis for creating logically complex queries, and a Guided Iterative Chain for robust CoT generation. Empirical validation on the BFCLv3 benchmark demonstrates that our approach achieves state-of-the-art performance among comparable-sized models. Furthermore, our model shows promising out-of-distribution generalization on BFCLv4, underscoring the robustness of our method. This research paves the way for developing more robust and stable models capable of complex reasoning and environment exploration.


\newpage
\bibliographystyle{config/antgroup}
\bibliography{references}

\begin{thebibliography}{14}
\providecommand{\natexlab}[1]{#1}
\providecommand{\url}[1]{\texttt{#1}}
\expandafter\ifx\csname urlstyle\endcsname\relax
  \providecommand{\doi}[1]{doi: #1}\else
  \providecommand{\doi}{doi: \begingroup \urlstyle{rm}\Url}\fi

\bibitem[Han et~al.(2024)Han, Zhang, Yao, Jin, and Xu]{han2024llm}
Shanshan Han, Qifan Zhang, Yuhang Yao, Weizhao Jin, and Zhaozhuo Xu.
\newblock Llm multi-agent systems: Challenges and open problems.
\newblock \emph{arXiv preprint arXiv:2402.03578}, 2024.

\bibitem[Hao et~al.(2025{\natexlab{a}})Hao, Wang, Xu, Chen, Peng, Gu, and Zhuang]{hao2025exploring}
Bingguang Hao, Maolin Wang, Zengzhuang Xu, Yicheng Chen, Cunyin Peng, Jinjie Gu, and Chenyi Zhuang.
\newblock Exploring superior function calls via reinforcement learning.
\newblock \emph{arXiv preprint arXiv:2508.05118}, 2025{\natexlab{a}}.

\bibitem[Hao et~al.(2025{\natexlab{b}})Hao, Wang, Xu, Peng, Chen, Zhao, Gu, and Zhuang]{hao2025funreason}
Bingguang Hao, Maolin Wang, Zengzhuang Xu, Cunyin Peng, Yicheng Chen, Xiangyu Zhao, Jinjie Gu, and Chenyi Zhuang.
\newblock Funreason: Enhancing large language models' function calling via self-refinement multiscale loss and automated data refinement.
\newblock \emph{arXiv preprint arXiv:2505.20192}, 2025{\natexlab{b}}.

\bibitem[Huang et~al.(2025)Huang, Gao, Shi, Wang, and Shang]{huang2025ttpa}
Chengrui Huang, Shen Gao, Zhengliang Shi, Dongsheng Wang, and Shuo Shang.
\newblock Ttpa: Token-level tool-use preference alignment training framework with fine-grained evaluation.
\newblock \emph{arXiv preprint arXiv:2505.20016}, 2025.

\bibitem[Ke et~al.(2025)Ke, Jiao, Ming, Nguyen, Xu, Long, Li, Qin, Wang, Savarese, et~al.]{ke2025survey}
Zixuan Ke, Fangkai Jiao, Yifei Ming, Xuan-Phi Nguyen, Austin Xu, Do~Xuan Long, Minzhi Li, Chengwei Qin, Peifeng Wang, Silvio Savarese, et~al.
\newblock A survey of frontiers in llm reasoning: Inference scaling, learning to reason, and agentic systems.
\newblock \emph{arXiv preprint arXiv:2504.09037}, 2025.

\bibitem[Liu et~al.(2024)Liu, Hoang, Zhang, Zhu, Lan, Tan, Yao, Liu, Feng, RN, et~al.]{liu2024apigen}
Zuxin Liu, Thai Hoang, Jianguo Zhang, Ming Zhu, Tian Lan, Juntao Tan, Weiran Yao, Zhiwei Liu, Yihao Feng, Rithesh RN, et~al.
\newblock Apigen: Automated pipeline for generating verifiable and diverse function-calling datasets.
\newblock \emph{Advances in Neural Information Processing Systems}, 37:\penalty0 54463--54482, 2024.

\bibitem[Patil et~al.()Patil, Mao, Yan, Ji, Suresh, Stoica, and Gonzalez]{patilberkeley}
Shishir~G Patil, Huanzhi Mao, Fanjia Yan, Charlie Cheng-Jie Ji, Vishnu Suresh, Ion Stoica, and Joseph~E Gonzalez.
\newblock The berkeley function calling leaderboard (bfcl): From tool use to agentic evaluation of large language models.
\newblock In \emph{Forty-second International Conference on Machine Learning}.

\bibitem[Prabhakar et~al.(2025)Prabhakar, Liu, Zhu, Zhang, Awalgaonkar, Wang, Liu, Chen, Hoang, Niebles, et~al.]{prabhakar2025apigen}
Akshara Prabhakar, Zuxin Liu, Ming Zhu, Jianguo Zhang, Tulika Awalgaonkar, Shiyu Wang, Zhiwei Liu, Haolin Chen, Thai Hoang, Juan~Carlos Niebles, et~al.
\newblock Apigen-mt: Agentic pipeline for multi-turn data generation via simulated agent-human interplay.
\newblock \emph{arXiv preprint arXiv:2504.03601}, 2025.

\bibitem[Sheng et~al.(2024)Sheng, Zhang, Ye, Wu, Zhang, Zhang, Peng, Lin, and Wu]{sheng2024hybridflow}
Guangming Sheng, Chi Zhang, Zilingfeng Ye, Xibin Wu, Wang Zhang, Ru~Zhang, Yanghua Peng, Haibin Lin, and Chuan Wu.
\newblock Hybridflow: A flexible and efficient rlhf framework.
\newblock \emph{arXiv preprint arXiv: 2409.19256}, 2024.

\bibitem[WANG et~al.(2025)WANG, ZHANG, PENG, CHEN, ZHOU, GU, ZHUANG, GUO, YU, WANG, et~al.]{wang2025function}
MAOLIN WANG, YINGYI ZHANG, CUNYIN PENG, YICHENG CHEN, WEI ZHOU, JINJIE GU, CHENYI ZHUANG, RUOCHENG GUO, BOWEN YU, WANYU WANG, et~al.
\newblock Function calling in large language models: Industrial practices, challenges, and future directions.
\newblock 2025.

\bibitem[Yang et~al.(2025)Yang, Li, Yang, Zhang, Hui, Zheng, Yu, Gao, Huang, Lv, et~al.]{yang2025qwen3}
An~Yang, Anfeng Li, Baosong Yang, Beichen Zhang, Binyuan Hui, Bo~Zheng, Bowen Yu, Chang Gao, Chengen Huang, Chenxu Lv, et~al.
\newblock Qwen3 technical report.
\newblock \emph{arXiv preprint arXiv:2505.09388}, 2025.

\bibitem[Zeng et~al.(2025)Zeng, Liu, Wang, Li, Mi, Wang, Shang, Jiang, and Liu]{zeng2025toolace}
Xingshan Zeng, Weiwen Liu, Lingzhi Wang, Liangyou Li, Fei Mi, Yasheng Wang, Lifeng Shang, Xin Jiang, and Qun Liu.
\newblock Toolace-mt: Non-autoregressive generation for agentic multi-turn interaction.
\newblock \emph{arXiv preprint arXiv:2508.12685}, 2025.

\bibitem[Zhang et~al.(2025)Zhang, Geng, Yu, Yin, Zhang, Tan, Zhou, Li, Xue, Li, et~al.]{zhang2025landscape}
Guibin Zhang, Hejia Geng, Xiaohang Yu, Zhenfei Yin, Zaibin Zhang, Zelin Tan, Heng Zhou, Zhongzhi Li, Xiangyuan Xue, Yijiang Li, et~al.
\newblock The landscape of agentic reinforcement learning for llms: A survey.
\newblock \emph{arXiv preprint arXiv:2509.02547}, 2025.

\bibitem[Zheng et~al.(2024)Zheng, Zhang, Zhang, Ye, Luo, Feng, and Ma]{zheng2024llamafactory}
Yaowei Zheng, Richong Zhang, Junhao Zhang, Yanhan Ye, Zheyan Luo, Zhangchi Feng, and Yongqiang Ma.
\newblock Llamafactory: Unified efficient fine-tuning of 100+ language models.
\newblock \emph{arXiv preprint arXiv:2403.13372}, 2024.

\end{thebibliography}

\newpage
\appendix

\section{Supplementary Case}
\label{app:additional_results}

This Section details the case study for FunReason-MT. We describe the synthesis process and a complete trajectory.

\definecolor{thinkcolor}{RGB}{138, 43, 226}    
\definecolor{toolcolor}{RGB}{34, 139, 34}      
\definecolor{toolresponsecolor}{RGB}{70, 130, 180}    
\definecolor{answercolor}{RGB}{25, 25, 112}    
\definecolor{usercolor}{RGB}{105, 105, 105}    
\definecolor{groundtruthcolor}{RGB}{139, 69, 19}    

\newcommand{\thinking}[1]{\textcolor{thinkcolor}{\textbf{#1}}}
\newcommand{\toolcall}[1]{\textcolor{toolcolor}{\textbf{#1}}}
\newcommand{\toolresponse}[1]{\textcolor{toolresponsecolor}{\textbf{#1}}}
\newcommand{\finalanswer}[1]{\textcolor{answercolor}{\textbf{#1}}}
\newcommand{\userquery}[1]{\textcolor{usercolor}{\textbf{#1}}}
\newcommand{\groundtruth}[1]{\textcolor{groundtruthcolor}{\textbf{#1}}}

\newtcolorbox{goodcasebox}[1][]{
    colback=green!8,
    colframe=green!50,
    boxrule=1.5pt,
    arc=3pt,
    left=8pt,
    right=8pt,
    top=6pt,
    bottom=6pt,
    breakable,
    #1
}

\newtcolorbox{badcasebox}[1][]{
    colback=red!8,
    colframe=red!50,
    boxrule=1.5pt,
    arc=3pt,
    left=8pt,
    right=8pt,
    top=6pt,
    bottom=6pt,
    breakable,
    #1
}

\newtcolorbox{neutralcasebox}[1][]{
    colback=gray!5,
    colframe=gray!40,
    boxrule=1pt,
    arc=3pt,
    left=8pt,
    right=8pt,
    top=6pt,
    bottom=6pt,
    breakable,
    #1
}

\newtcolorbox{querybox}{
    colback=usercolor!10,
    colframe=usercolor!60,
    boxrule=1.5pt,
    arc=4pt,
    left=10pt,
    right=10pt,
    top=8pt,
    bottom=8pt,
    fontupper=\small
}

\newtcolorbox{groundtruthbox}{
    colback=groundtruthcolor!10,
    colframe=groundtruthcolor!60,
    boxrule=1.5pt,
    arc=4pt,
    left=10pt,
    right=10pt,
    top=8pt,
    bottom=8pt,
    fontupper=\small
}

\newtcolorbox{trajectorybox}{
    colback=white,
    colframe=gray!30,
    boxrule=1pt,
    arc=2pt,
    left=6pt,
    right=6pt,
    top=4pt,
    bottom=4pt,
    breakable,
    fontupper=\footnotesize
}

\newtcolorbox{toolresponsebox}{
    colback=toolresponsecolor!10,
    colframe=toolresponsecolor!60,
    boxrule=1pt,
    arc=2pt,
    left=6pt,
    right=6pt,
    top=4pt,
    bottom=4pt,
    breakable,
    fontupper=\footnotesize
}

\begin{toolresponsebox}
    
\textbf{Case:} 
We showcase the construction of Hard Query with logical jump that directly enables ticket purchase, which involves a jump for consulting the city zip code.
\begin{querybox}
\textbf{Trace:}

get\_zipcode(city="Rivermist")

output: "83214"

get\_zipcode(city="Stonebrook")

output: "74532"

buy\_tickets(zipcodeA="83214", zipcodeB="74532")

output: "ticket\_id: 14589"
   
\textbf{Advanced Tool:} 

buy\_tickets\_adv(input\_param: cityA, cityB, 
output: ticket\_id, desc: purchase air tickets between two cities by city names, returning the purchased ticket information.)

\textbf{Easy Query: }\\
Please check the zip code of the city "Rivermist" and "Stonebrook" first, then purchase air tickets between the two cities according to the zip codes you checked.

\textbf{Hard Query: }\\
Please purchase air tickets between the city "Rivermist" and the city "Stonebrook".

\textbf{Hint:}

"purchase air tickets between two cities by city names, returning the purchased ticket information" is equal to "retrieve the zipcode for a given city" and "buy tickets from cityA to cityB with zipcode."

\textbf{Assistant:}

\thinking{<think>}

Okay, so the user wants me to purchase air tickets between Rivermist and Stonebrook. Let me see how I can approach this using the tools provided. First, I need to understand the tools available. There's `get\_zipcode` which retrieves a zipcode for a given city, and `buy\_tickets` which buys tickets using the zipcodes of the start and target cities. Hmm, the user's query is about purchasing tickets between two cities, but the tools require zipcodes. So, I can't directly buy tickets with city names; I need to get their zipcodes first. Alright, step one: use `get\_zipcode` for Rivermist. I'll call it with the city parameter as "Rivermist". This should give me the zipcode for Rivermist. Next, I'll do the same for Stonebrook. Another call to `get\_zipcode` with "Stonebrook" as the city parameter. That should give me the zipcode for Stonebrook. Once I have both zipcodes, I can use the `buy\_tickets` tool. I'll input the zipcodes of Rivermist and Stonebrook into `buy\_tickets` to purchase the tickets. Wait, the output format the user provided is a list of function calls in a specific format. So I need to structure my response with the function names and their parameters correctly. I should make sure that each function call is properly formatted, especially the parameters. Since the city names are strings, I'll enclose them in quotes as per the user's instructions. Let me double-check the parameters for each function. For `get\_zipcode`, it's just the city name. For `buy\_tickets`, it's both cityA\_zipcode and cityB\_zipcode. I need to ensure that I pass these correctly in the order specified. I think I've got the sequence right. First, get the zipcodes for both cities, then use those zipcodes to buy the tickets. That way, the tools are used in the correct order to achieve the user's goal. I should also consider if there are any edge cases, like if a city doesn't have a zipcode. But since the user provided specific cities, I'll assume the tools can handle them. Alright, putting it all together, I'll list the function calls in the required format, making sure each parameter is correctly specified with quotes around the string values.

\thinking{</think>}

        \toolcall{<tool\_call>}

[get\_zipcode(city="Rivermist"), get\_zipcode(city="Stonebrook"), buy\_tickets(cityA\_zipcode="12345", cityB\_zipcode="67890")]

        \toolcall{</tool\_call>}
    \end{querybox}
\begin{querybox}
\groundtruth{Turn:}

\textbf{Query:}

Please purchase air tickets between the city "Rivermist" and the city "Stonebrook".

\textbf{Tools:}

get\_zipcode, buy\_ticket, $\dots$

\textbf{Assistant:}

\thinking{<think>}

Okay, so the user wants me to purchase air tickets between Rivermist and Stonebrook. Let me see how I can approach this using the tools provided. First, I need to understand the tools available. There's `get\_zipcode` which retrieves a zipcode for a given city, and `buy\_tickets` which buys tickets using the zipcodes of the start and target cities. Hmm, the user's query is about purchasing tickets between two cities, but the tools require zipcodes. So, I can't directly buy tickets with city names; I need to get their zipcodes first. Alright, step one: use `get\_zipcode` for Rivermist. I'll call it with the city parameter as "Rivermist". This should give me the zipcode for Rivermist. Next, I'll do the same for Stonebrook. Another call to `get\_zipcode` with "Stonebrook" as the city parameter. That should give me the zipcode for Stonebrook. Once I have both zipcodes, I can use the `buy\_tickets` tool. I'll input the zipcodes of Rivermist and Stonebrook into `buy\_tickets` to purchase the tickets. Wait, the output format the user provided is a list of function calls in a specific format. So I need to structure my response with the function names and their parameters correctly. I should make sure that each function call is properly formatted, especially the parameters. Since the city names are strings, I'll enclose them in quotes as per the user's instructions. Let me double-check the parameters for each function. For `get\_zipcode`, it's just the city name. For `buy\_tickets`, it's both cityA\_zipcode and cityB\_zipcode. I need to ensure that I pass these correctly in the order specified. I think I've got the sequence right. First, get the zipcodes for both cities, then use those zipcodes to buy the tickets. That way, the tools are used in the correct order to achieve the user's goal. I should also consider if there are any edge cases, like if a city doesn't have a zipcode. But since the user provided specific cities, I'll assume the tools can handle them. Alright, putting it all together, I'll list the function calls in the required format, making sure each parameter is correctly specified with quotes around the string values.

\thinking{</think>}

        \toolcall{<tool\_call>}

[get\_zipcode(city="Rivermist"), get\_zipcode(city="Stonebrook"), buy\_tickets(cityA\_zipcode="12345", cityB\_zipcode="67890")]

\toolcall{</tool\_call>}

\end{querybox}
\end{toolresponsebox}

\newpage

\begin{toolresponsebox}
    \textbf{Trajectory:} We showcase a two-turn complete trajectory.


        \textbf{User:} 
        
        Determine the current working directory path, and then search for all files and directories in the current directory and its subdirectories.

    \textbf{Assistant:}

        \thinking{<think>}

Okay, let's tackle the user's request step by step. The user wants to determine the current working directory path and then search for all files and directories in the current directory and its subdirectories.

First, I need to figure out the current working directory. The 'pwd' command is perfect for that. It doesn't require any parameters, so I'll call it first. That should give the user the path they're in.

Next, the user wants to search all files and directories in the current directory and subdirectories. The 'find' function can do this. The 'find' tool has parameters 'path' and 'name'. The default for 'path' is ".", which is the current directory. Since they want everything, the 'name' parameter should be None (or omitted) to return all items. The default for 'name' is already None, so I just need to call find with the default parameters. 

Wait, the user specified "its subdirectories" so the recursive search is covered by the find tool's description. So using find with path "." and no name should work. I don't need to specify any parameters except maybe path, but since it's default, it's okay. 

So the function calls should be first pwd, then find. Let me check the parameters again. The find function's 'path' defaults to ".", which is correct. The 'name' is optional and defaults to None. So the correct calls are [pwd(), find(path=.)]. 

I need to make sure that the parameters are correctly formatted. Since 'path' is set to ".", even though it's the default, including it explicitly might be better to ensure clarity. The user might expect the search to start at the current directory, so specifying path="." is safe. 

No other functions are needed here. The 'ls' command only lists the current directory, not subdirectories, so 'find' is the right choice. Alright, that should cover the user's needs.

        \thinking{</think>}

        \toolcall{<tool\_call>}

        [pwd(), find(path=".")]

        \toolcall{</tool\_call>}

\textbf{Tool:}

        \toolresponse{<tool\_response>}

               [{'pwd()': '{"current\_working\_directory": "/workspace"}'}, {'find(path=".")': '{"matches": ["./processed", "./processed/vqrqs3933.md", "./processed/akqff.md", "./processed/documents", "./requirements.txt", "./dqyiu\_68828.json", "./ulyuo.json", "./data", "./data/raw", "./data/raw/dtpsp.json", "./data/metadata.json", "./data/vwccl.md", "./data/unggv.json", "./data/code", "./data/project"]}'}]

        \toolresponse{</tool\_response>}

\textbf{Assistant:}
    
        \thinking{<think>}

        Okay, let's see what the user needs here. The user asked to determine the current working directory and then search all files and directories in the current directory and subdirectories.
        
        First, the assistant already called pwd() and find(path=".") in the previous turn. The tool responses show that pwd() returned "/workspace" as the current directory. The find function with path "." gave a list of matches including various files and directories under the current directory and its subdirectories. 
        
        Now, the user's requirement is to address their query in the next step. Since the user wanted both the current directory and a search, the assistant already executed both functions successfully. The tool\_response includes the path and the matches. 
        
        The thought process should confirm that both functions were executed and their results are available. The reply should describe the results. Since the user's needs were met, there's no need for further function calls. The assistant should present the findings, like stating the current directory and listing the found items. But according to the output format, the reply\_to\_the\_user should either be function calls or a description. Since the functions are already done, the reply should be a description summarizing the results. 
        
        Wait, the output format says if you need to give a description, output a sentence. So the assistant should say something like "The current working directory is /workspace. The search found the following items: [list of matches]." But the matches are in the tool\_response's find result. However, the user might just need the info presented. 
        
        But looking back at the instructions, in the reply\_to\_the\_user section, if you need to give a description, output a sentence. Since the functions have been executed, the next step is to describe the results. So the assistant should format the answer with the current directory and the matches from find. 
        
        Wait, but the example given in the instructions for reply\_to\_the\_user when giving a description is like "the cd function is executed success, and current workdirctory is src." So in this case, the assistant should combine both results. 
        
        So the thought process would note that both functions were executed successfully. The reply should state the current directory and list the found items. But since the matches are many, maybe just mention that the search found items like those listed. 
        
        Alternatively, the exact answer should reference the tool\_response data. The current\_working\_directory is /workspace, and the matches array includes all those paths. So the assistant can present that info concisely.

        \thinking{</think>}

        \toolcall{<tool\_call>}

        The current working directory is /workspace. The search found the following items: ["./processed", "./processed/vqrqs3933.md", "./processed/akqff.md", "./processed/documents", "./requirements.txt", "./dqyiu\_68828.json", "./ulyuo.json", "./data", "./data/raw", "./data/raw/dtpsp.json", "./data/metadata.json", "./data/vwccl.md", "./data/unggv.json", "./data/code", "./data/project"]."

        \toolcall{</tool\_call>}

        \textbf{User:} 
        
        Display the contents of the file named unggv.json in the data directory, and then display the last 7 lines of the file named dtpsp.json, which is in the data directory of the raw directory.

\textbf{Assistant:}

        \thinking{<think>}

                    Okay, let's tackle the user's latest request. They want to display the contents of "unggv.json" in the data directory and the last 7 lines of "dtpsp.json" located in the data/raw directory.
                    
                    First, I need to check the current directory. From the previous tool response, the current working directory is "/workspace". The user mentioned "data directory", so "unggv.json" is directly under "/workspace/data/unggv.json". But to access it, I should use the 'cat' command. However, 'cat' requires the file to be in the current directory. Since the current directory is "/workspace", I need to check if "unggv.json" is there. Wait, looking back at the find results, "unggv.json" is listed under "./data/unggv.json". So it's in the data subdirectory. To access it, I need to either change into the data directory first or specify the path, but the 'cat' tool only allows the file name in the current directory. Therefore, I should first 'cd' into the data directory. 
                    
                    Wait, the 'cd' tool changes the directory, so first step: cd into 'data'. Then, use 'cat unggv.json'. After that, for the second part, the user wants the last 7 lines of 'dtpsp.json' in data/raw. Since we are now in the data directory, the 'raw' directory is a subdirectory. So we need to 'cd raw' to get into that directory, then use 'tail' with lines=7 on 'dtpsp.json'. But after that, we might need to go back, but the user didn't ask for that. Alternatively, maybe we can do it without changing directories again by specifying the path, but according to the tool's parameters, 'tail' only takes the file name in the current directory. 
                    
                    Wait, the 'tail' parameters say "No path is allowed and you can only perform on file at local directory." So to access 'data/raw/dtpsp.json', we need to be in the 'data/raw' directory. So the steps would be:
                    
                    1. Current directory is /workspace. First, to access 'unggv.json' in data/, do 'cd data', then 'cat unggv.json'.
                    2. Then, to get into data/raw, do 'cd raw', then 'tail(file\_name="dtpsp.json", lines=7)'.
                    
                    But after the first 'cd data', the current directory becomes /workspace/data. Then after 'cd raw', it becomes /workspace/data/raw. Then after those commands, the user might want to return, but since it's a multi-step process, the assistant should just proceed step by step.
                    
                    However, the user's request is to first display unggv.json and then dtpsp.json. So the function calls would be:
                    
                    First, cd into data: [cd(folder="data")]
                    
                    Then, cat unggv.json: [cat(file\_name="unggv.json")]
                    
                    Then, cd into raw: [cd(folder="raw")]
                    
                    Then, tail dtpsp.json with 7 lines: [tail(file\_name="dtpsp.json", lines=7)]
                    
                    But after these steps, the current directory is now data/raw. The user might need to go back, but since the user didn't ask, maybe it's okay. Alternatively, maybe there's another way without changing directories multiple times, but according to the tool constraints, it's necessary.
                    
                    Wait, but the 'find' result shows that 'dtpsp.json' is in "./data/raw/dtpsp.json". So the path from the current directory (which is /workspace) would require navigating into data/raw. So the steps are necessary.
                    
                    Therefore, the function calls should be in the order:
                    
                    1. cd into data
                    2. cat unggv.json
                    3. cd into raw
                    4. tail dtpsp.json with lines=7
                    
                    So the function calls would be:
                    
                    [cd(folder="data"), cat(file\_name="unggv.json"), cd(folder="raw"), tail(file\_name="dtpsp.json", lines=7)]
                    
                    But I need to make sure each step is correct. Let me check each function's parameters again.
                    
                    For 'cd', the parameter is 'folder', which is the directory to change to. So first cd to "data", then from there cd to "raw" (since current is now data/). 
                    
                    The 'cat' requires the file name in current directory. After cd data, unggv.json is there. Then, after cd raw, dtpsp.json is in current directory. So yes, that works.
                    
                    Therefore, the correct function calls are as above.
                    
        \thinking{</think>}

        \toolcall{<tool\_call>}

        [cd(folder="data"), cat(file\_name="unggv.json"), cd(folder="raw"), tail(file\_name="dtpsp.json", lines=7)]

        \toolcall{</tool\_call>}

        \textbf{Tool:}

        \toolresponse{<tool\_response>}

        [{'cd(folder="data")': '{"current\_working\_directory": "data"}'}, {'cat(file\_name="unggv.json")': '{"file\_content": "{  "data": { "id": "94",    "type": "product"  }}"}'}, {'cd(folder="raw")': '{"current\_working\_directory": "raw"}'}, {'tail(file\_name="dtpsp.json", lines=7)': '{"last\_lines": "{  "data": {    "id": "27",    "type": "user"  }}"}'}]
        
        \toolresponse{</tool\_response>}

\textbf{Assistant:}
    
        \thinking{<think>}

        Okay, let's see what the user is asking for now. The user's latest query was to display the contents of "unggv.json" in the data directory and the last 7 lines of "dtpsp.json" located in the data/raw directory.
        
        Looking at the previous interactions, the assistant first used pwd() and find() to get the current directory and list all files. Then, the user's next request required navigating into the data directory with cd("data"), then using cat on unggv.json. After that, they changed to the raw directory with cd("raw") and used tail on dtpsp.json with lines=7. 
        
        The tool responses show that all these commands were executed successfully. The current working directory was first changed to "data", then to "raw". The cat and tail commands both returned the expected file contents. Since all the functions called in the assistant's response have been executed without errors, there's no need to repeat any commands. The user's requirements have been fully met. Now, the assistant should present the results of these commands to the user in a clear manner, mentioning the contents of unggv.json and the last lines of dtpsp.json as per the tool responses.

        \thinking{</think>}

        \finalanswer{<answer>}

        The contents of "unggv.json" in the data directory are:  
        "{  "data": {    "id": "94",    "type": "product"  }}".  
        
        The last 7 lines of "dtpsp.json" in the data/raw directory are:  
        "{  "data": {    "id": "27",    "type": "user"  }}"."

        \finalanswer{</answer>}

\end{toolresponsebox}

\end{document}